\documentclass[10pt,twocolumn,letterpaper]{article}

\usepackage{cvpr}
\usepackage{times}
\usepackage{epsfig}
\usepackage{graphicx}
\usepackage{amsmath}
\usepackage{amssymb}
\usepackage{epstopdf}
\usepackage{booktabs}

\usepackage[pagebackref=true,breaklinks=true,letterpaper=true,colorlinks,bookmarks=false]{hyperref}

\cvprfinalcopy 

\def\cvprPaperID{1054} 

\ifcvprfinal\fi
\begin{document}

\title{Pyramid Scene Parsing Network}


\author{
Hengshuang Zhao$^{1}$ \quad Jianping Shi$^{2}$ \quad Xiaojuan Qi$^{1}$ \quad Xiaogang Wang$^{1}$ \quad Jiaya Jia$^{1}$\\
$^{1}$The Chinese University of Hong Kong \quad $^{2}$SenseTime Group Limited\\
{\tt\small \{hszhao, xjqi, leojia\}@cse.cuhk.edu.hk, xgwang@ee.cuhk.edu.hk, shijianping@sensetime.com}}

\maketitle
\begin{abstract}
Scene parsing is challenging for unrestricted open vocabulary and diverse scenes. In this
paper, we exploit the capability of global context information by different-region-based
context aggregation through our pyramid pooling module together with the proposed pyramid
scene parsing network (PSPNet). Our global prior representation is effective to produce
good quality results on the scene parsing task, while PSPNet provides a superior
framework for pixel-level prediction. The proposed approach achieves state-of-the-art
performance on various datasets. It came first in ImageNet scene parsing challenge 2016,
PASCAL VOC 2012 benchmark and Cityscapes benchmark. A single PSPNet yields the new record
of mIoU accuracy 85.4\% on PASCAL VOC 2012 and accuracy 80.2\% on Cityscapes.
\end{abstract}


\section{Introduction}

Scene parsing, based on semantic segmentation, is a fundamental topic in computer vision.
The goal is to assign each pixel in the image a category label. Scene parsing provides
complete understanding of the scene. It predicts the label, location, as well as shape
for each element. This topic is of broad interest for potential applications of automatic
driving, robot sensing, to name a few.

\begin{figure}[t]
\centering
\includegraphics[width=1.0\linewidth]{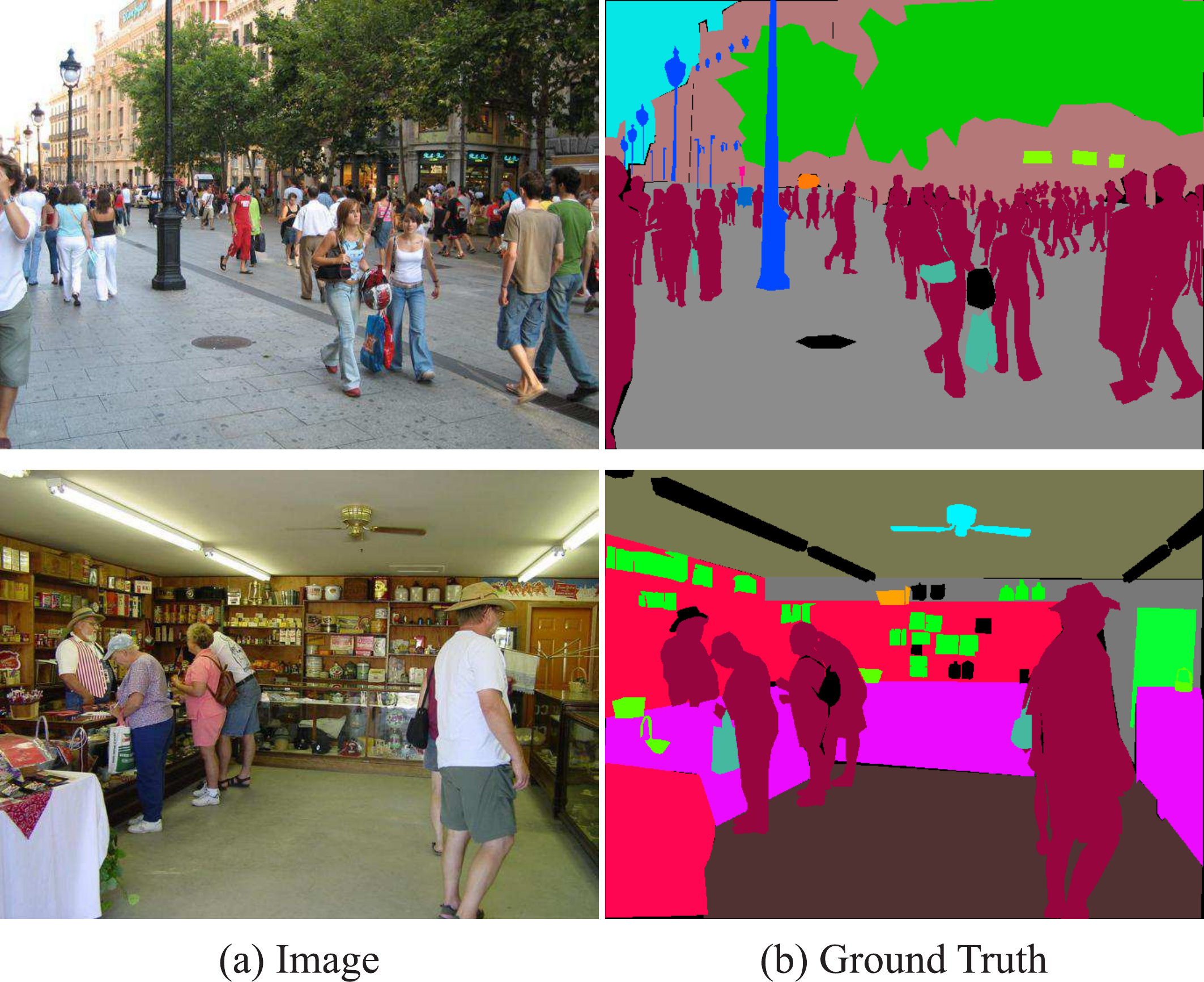}
\caption{Illustration of complex scenes in ADE20K dataset. } \label{fig:intro}
\end{figure}

Difficulty of scene parsing is closely related to scene and label variety. The pioneer
scene parsing task~\cite{liu2011nonparametric} is to classify 33 scenes for 2,688 images
on LMO dataset~\cite{liu2009nonparametric}. More recent PASCAL VOC semantic segmentation
and PASCAL context datasets~\cite{everingham2010pascal,mottaghi2014role} include more
labels with similar context, such as chair and sofa, horse and cow, etc. The new ADE20K
dataset~\cite{zhou2016semantic} is the most challenging one with a large and unrestricted
open vocabulary and more scene classes. A few representative images are shown in
Fig.~\ref{fig:intro}. To develop an effective algorithm for these datasets needs to
conquer a few difficulties.

\begin{figure*}
\begin{center}
\includegraphics[width=0.99\linewidth]{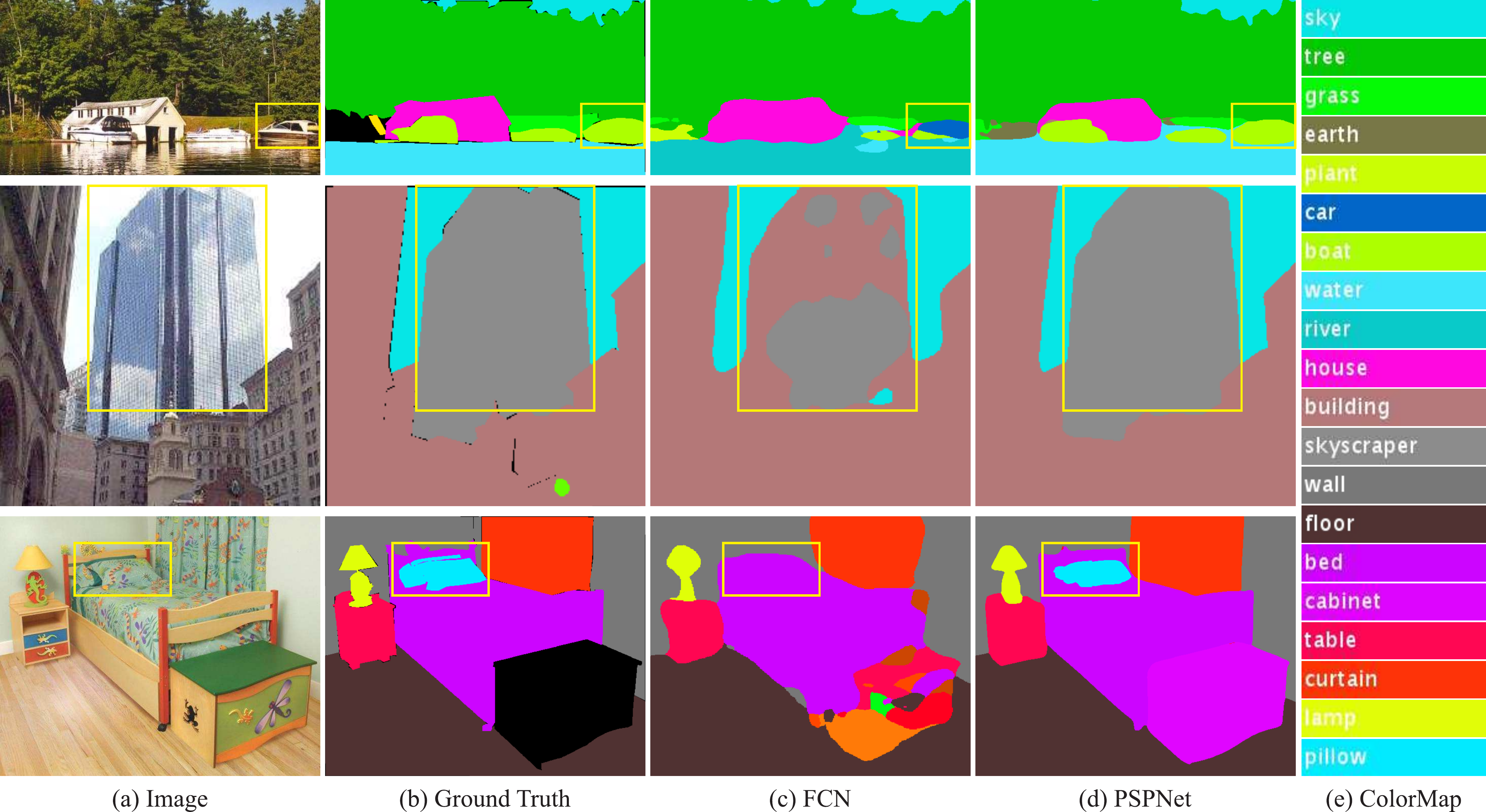}
\end{center}
\caption{Scene parsing issues we observe on ADE20K~\cite{zhou2016semantic} dataset. The
first row shows the issue of mismatched relationship -- cars are seldom over water than
boats. The second row shows confusion categories where class ``building" is easily
confused as ``skyscraper". The third row illustrates inconspicuous classes. In this
example, the pillow is very similar to the bed sheet in terms of color and texture. These
inconspicuous objects are easily misclassified by FCN. } \label{fig:fcnissues}
\end{figure*}

State-of-the-art scene parsing frameworks are mostly based on the {\it fully
convolutional network} (FCN)~\cite{long2015fully}. The deep {\it convolutional neural
network} (CNN) based methods boost dynamic object understanding, and yet still face
challenges considering diverse scenes and unrestricted vocabulary. One example is shown
in the first row of Fig.~\ref{fig:fcnissues}, where a {\it boat} is mistaken as a {\it
car}. These errors are due to similar appearance of objects. But when viewing the image
regarding the context prior that the scene is described as {\it boathouse} near a river,
correct prediction should be yielded.

Towards accurate scene perception, the knowledge graph relies on prior information of
scene context. We found that the major issue for current FCN based models is lack of
suitable strategy to utilize global scene category clues. For typical complex scene
understanding, previously to get a global image-level feature, spatial pyramid
pooling~\cite{lazebnik2006beyond} was widely employed where spatial statistics provide a
good descriptor for overall scene interpretation. Spatial pyramid pooling
network~\cite{he2014spatial} further enhances the ability.

Different from these methods, to incorporate suitable global features, we propose {\it
pyramid scene parsing network} (PSPNet). In addition to traditional dilated FCN
\cite{chen2014semantic,yu2015multi} for pixel prediction, we extend the pixel-level
feature to the specially designed global pyramid pooling one. The local and global clues
together make the final prediction more reliable. We also propose an optimization
strategy with deeply supervised loss. We give all implementation details, which are key
to our decent performance in this paper, and make the code and trained models publicly
available
\footnote{\href{https://github.com/hszhao/PSPNet}{https://github.com/hszhao/PSPNet}}.

Our approach achieves state-of-the-art performance on all available datasets. It is the
champion of ImageNet scene parsing challenge 2016~\cite{zhou2016semantic}, and arrived
the 1st place on PASCAL VOC 2012 semantic segmentation
benchmark~\cite{everingham2010pascal}, and the 1st place on urban scene Cityscapes
data~\cite{cordts2016cityscapes}. They manifest that PSPNet gives a promising direction
for pixel-level prediction tasks, which may even benefit CNN-based stereo matching,
optical flow, depth estimation, etc. in follow-up work. Our main contributions are
threefold.
\begin{itemize}
\vspace{-0.1cm}
\item
We propose a pyramid scene parsing network to embed difficult scenery context features in
an FCN based pixel prediction framework. \vspace{-0.1cm}
\item
We develop an effective optimization strategy for deep ResNet~\cite{he2015deep} based on
deeply supervised loss. \vspace{-0.1cm}
\item
We build a practical system for state-of-the-art scene parsing and semantic segmentation
where all crucial implementation details are included.
\end{itemize}

\section{Related Work}

In the following, we review recent advances in scene parsing and semantic segmentation
tasks. Driven by powerful deep neural
networks~\cite{krizhevsky2012imagenet,simonyan2014very,szegedy2015going,he2015deep},
pixel-level prediction tasks like scene parsing and semantic segmentation achieve great
progress inspired by replacing the fully-connected layer in classification with the
convolution layer~\cite{long2015fully}. To enlarge the receptive field of neural
networks, methods of~\cite{chen2014semantic,yu2015multi} used dilated convolution.
Noh~\etal~\cite{noh2015learning} proposed a coarse-to-fine structure with deconvolution
network to learn the segmentation mask. Our baseline network is FCN and dilated
network~\cite{long2015fully,chen2014semantic}.

Other work mainly proceeds in two directions. One
line~\cite{long2015fully,chen2014semantic,chen2015attention,xia2016zoom,hariharan2015hypercolumns}
is with multi-scale feature ensembling. Since in deep networks, higher-layer feature
contains more semantic meaning and less location information. Combining multi-scale
features can improve the performance.

The other direction is based on structure prediction. The pioneer work
\cite{chen2014semantic} used conditional random field (CRF) as post processing to refine
the segmentation result. Following
methods~\cite{liu2015semantic,zheng2015conditional,arnab2016higher} refined networks via
end-to-end modeling. Both of the two directions ameliorate the localization ability of
scene parsing where predicted semantic boundary fits objects. Yet there is still much
room to exploit necessary information in complex scenes.

To make good use of global image-level priors for diverse scene understanding, methods of
\cite{lazebnik2006beyond,lucchi2011spatial} extracted global context information with
traditional features not from deep neural networks. Similar improvement was made under
object detection frameworks \cite{szegedy2014scalable}. Liu~\etal~\cite{liu2015parsenet}
proved that global average pooling with FCN can improve semantic segmentation results.
However, our experiments show that these global descriptors are not representative enough
for the challenging ADE20K data. Therefore, different from global pooling
in~\cite{liu2015parsenet}, we exploit the capability of global context information by
different-region-based context aggregation via our pyramid scene parsing network.

\section{Pyramid Scene Parsing Network}

\begin{figure*}
\begin{center}
\includegraphics[width=1.0\linewidth]{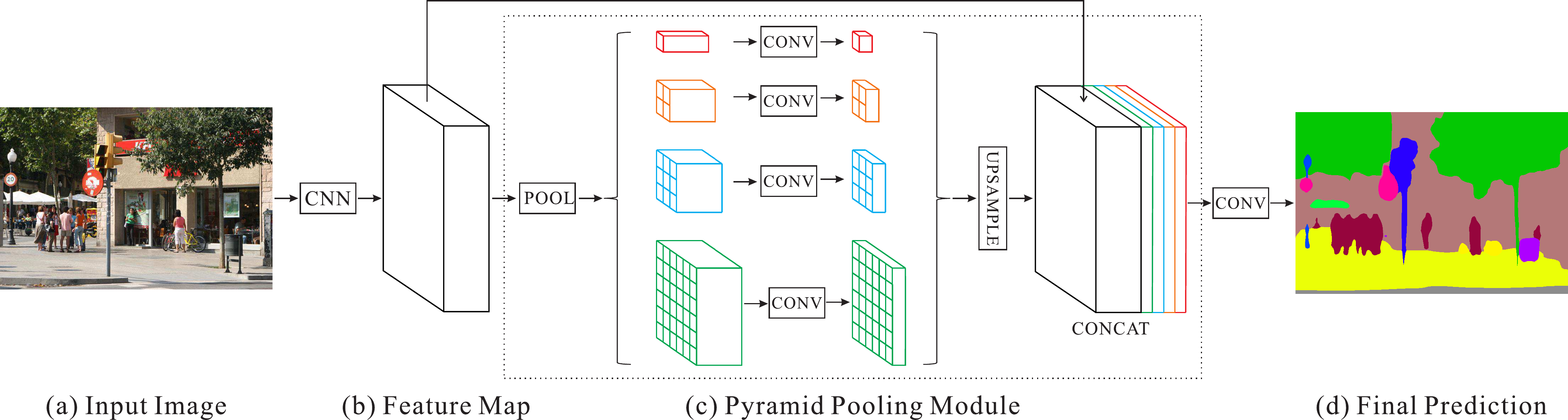}
\end{center}
\caption{Overview of our proposed PSPNet. Given an input image (a), we first use CNN to
get the feature map of the last convolutional layer (b), then a pyramid parsing module is
applied to harvest different sub-region representations, followed by upsampling and
concatenation layers to form the final feature representation, which carries both local
and global context information in (c). Finally, the representation is fed into a
convolution layer to get the final per-pixel prediction (d).} \label{fig:pspnet}
\end{figure*}

We start with our observation and analysis of representative failure cases when applying
FCN methods to scene parsing. They motivate proposal of our pyramid pooling module as the
effective global context prior. Our {\it pyramid scene parsing network} (PSPNet)
illustrated in Fig.~\ref{fig:pspnet} is then described to improve performance for
open-vocabulary object and stuff identification in complex scene parsing.

\subsection{Important Observations}

The new ADE20K dataset \cite{zhou2016semantic} contains 150 stuff/object category labels
(\eg, wall, sky, and tree) and 1,038 image-level scene descriptors (\eg,
airport\_terminal, bedroom, and street). So a large amount of labels and vast
distributions of scenes come into existence. Inspecting the prediction results of the FCN
baseline provided in~\cite{zhou2016semantic}, we summarize several common issues for
complex-scene parsing.

\vspace{-0.1in}
\paragraph{Mismatched Relationship} Context relationship is universal and
important especially for complex scene understanding. There exist co-occurrent visual
patterns. For example, an airplane is likely to be in runway or fly in sky while not over
a road. For the first-row example in Fig.~\ref{fig:fcnissues}, FCN predicts the boat in
the yellow box as a ``car" based on its appearance. But the common knowledge is that a
car is seldom over a river. Lack of the ability to collect contextual information
increases the chance of misclassification.

\vspace{-0.1in}
\paragraph{Confusion Categories}
There are many class label pairs in the ADE20K dataset~\cite{zhou2016semantic} that are
confusing in classification. Examples are {\it field} and {\it earth}; {\it mountain} and
{\it hill}; {\it wall}, {\it house}, {\it building} and {\it skyscraper}. They are with
similar appearance. The expert annotator who labeled the entire dataset, still makes
17.60\% pixel error as described in \cite{zhou2016semantic}. In the second row of
Fig.~\ref{fig:fcnissues}, FCN predicts the object in the box as part of {\it skyscraper}
and part of {\it building}. These results should be excluded so that the whole object is
either {\it skyscraper} or {\it building}, but not both. This problem can be remedied by
utilizing the relationship between categories.

\vspace{-0.1in}
\paragraph{Inconspicuous Classes}
Scene contains objects/stuff of arbitrary size. Several small-size things, like
streetlight and signboard, are hard to find while they may be of great importance.
Contrarily, big objects or stuff may exceed the receptive field of FCN and thus cause
discontinuous prediction. As shown in the third row of Fig.~\ref{fig:fcnissues}, the
pillow has similar appearance with the sheet. Overlooking the global scene category may
fail to parse the pillow. To improve performance for remarkably small or large objects,
one should pay much attention to different sub-regions that contain
inconspicuous-category stuff.

To summarize these observations, many errors are partially or completely related to
contextual relationship and global information for different receptive fields. Thus a
deep network with a suitable global-scene-level prior can much improve the performance of
scene parsing.

\subsection{Pyramid Pooling Module}

With above analysis, in what follows, we introduce the pyramid pooling module, which
empirically proves to be an effective global contextual prior.

In a deep neural network, the size of receptive field can roughly indicates how much we
use context information. Although theoretically the receptive field of
ResNet~\cite{he2015deep} is already larger than the input image, it is shown by
Zhou~\etal~\cite{zhou2014object} that the empirical receptive field of CNN is much
smaller than the theoretical one especially on high-level layers. This makes many
networks not sufficiently incorporate the momentous global scenery prior. We address this
issue by proposing an effective global prior representation.

Global average pooling is a good baseline model as the global contextual prior, which is
commonly used in image classification tasks~\cite{szegedy2015going,he2015deep}. In
\cite{liu2015parsenet}, it was successfully applied to semantic segmentation. But
regarding the complex-scene images in ADE20K~\cite{zhou2016semantic}, this strategy is
not enough to cover necessary information. Pixels in these scene images are annotated
regarding many stuff and objects. Directly fusing them to form a single vector may lose
the spatial relation and cause ambiguity. Global context information along with
sub-region context is helpful in this regard to distinguish among various categories. A
more powerful representation could be fused information from different sub-regions with
these receptive fields. Similar conclusion was drawn in classical work
\cite{lazebnik2006beyond,he2014spatial} of scene/image classification.

In \cite{he2014spatial}, feature maps in different levels generated by pyramid pooling
were finally flattened and concatenated to be fed into a fully connected layer for
classification. This global prior is designed to remove the fixed-size constraint of CNN
for image classification. To further reduce context information loss between different
sub-regions, we propose a hierarchical global prior, containing information with
different scales and varying among different sub-regions. We call it {\it pyramid pooling
module} for global scene prior construction upon the final-layer-feature-map of the deep
neural network, as illustrated in part (c) of Fig.~\ref{fig:pspnet}.

The pyramid pooling module fuses features under four different pyramid scales. The
coarsest level highlighted in red is global pooling to generate a single bin output. The
following pyramid level separates the feature map into different sub-regions and forms
pooled representation for different locations. The output of different levels in the
pyramid pooling module contains the feature map with varied sizes. To maintain the weight
of global feature, we use $1\times 1$ convolution layer after each pyramid level to
reduce the dimension of context representation to $1/N$ of the original one if the level
size of pyramid is $N$. Then we directly upsample the low-dimension feature maps to
get the same size feature as the original feature map via bilinear
interpolation. Finally, different levels of features are concatenated as the final
pyramid pooling global feature.

Noted that the number of pyramid levels and size of each level can be modified. They are
related to the size of feature map that is fed into the pyramid pooling layer. The
structure abstracts different sub-regions by adopting varying-size pooling kernels in a
few strides. Thus the multi-stage kernels should maintain a reasonable gap in
representation. Our pyramid pooling module is a four-level one with bin sizes of $1\times
1$, $2\times 2$, $3\times 3$ and $6\times 6$ respectively. For the type of pooling
operation between max and average, we perform extensive experiments to show the
difference in Section \ref{sec:imagenetexperimet}.

\subsection{Network Architecture}

With the pyramid pooling module, we propose our {\it pyramid scene parsing} network
(PSPNet) as illustrated in Fig.~\ref{fig:pspnet}. Given an input image in
Fig.~\ref{fig:pspnet}(a), we use a pretrained ResNet~\cite{he2015deep} model with the
dilated network strategy~\cite{chen2014semantic,yu2015multi} to extract the feature map.
The final feature map size is $1/8$ of the input image, as shown in
Fig.~\ref{fig:pspnet}(b). On top of the map, we use the pyramid pooling module shown in
(c) to gather context information. Using our 4-level pyramid, the pooling kernels cover
the whole, half of, and small portions of the image. They are fused as the global prior.
Then we concatenate the prior with the original feature map in the final part of (c). It
is followed by a convolution layer to generate the final prediction map in (d).

To explain our structure, PSPNet provides an effective global contextual prior for
pixel-level scene parsing. The pyramid pooling module can collect levels of information,
more representative than global pooling \cite{liu2015parsenet}. In terms of computational
cost, our PSPNet does not much increase it compared to the original dilated FCN network.
In end-to-end learning, the global pyramid pooling module and the local FCN feature can
be optimized simultaneously.

\begin{figure}[t]
\begin{center}
    \includegraphics[width=1.0\linewidth]{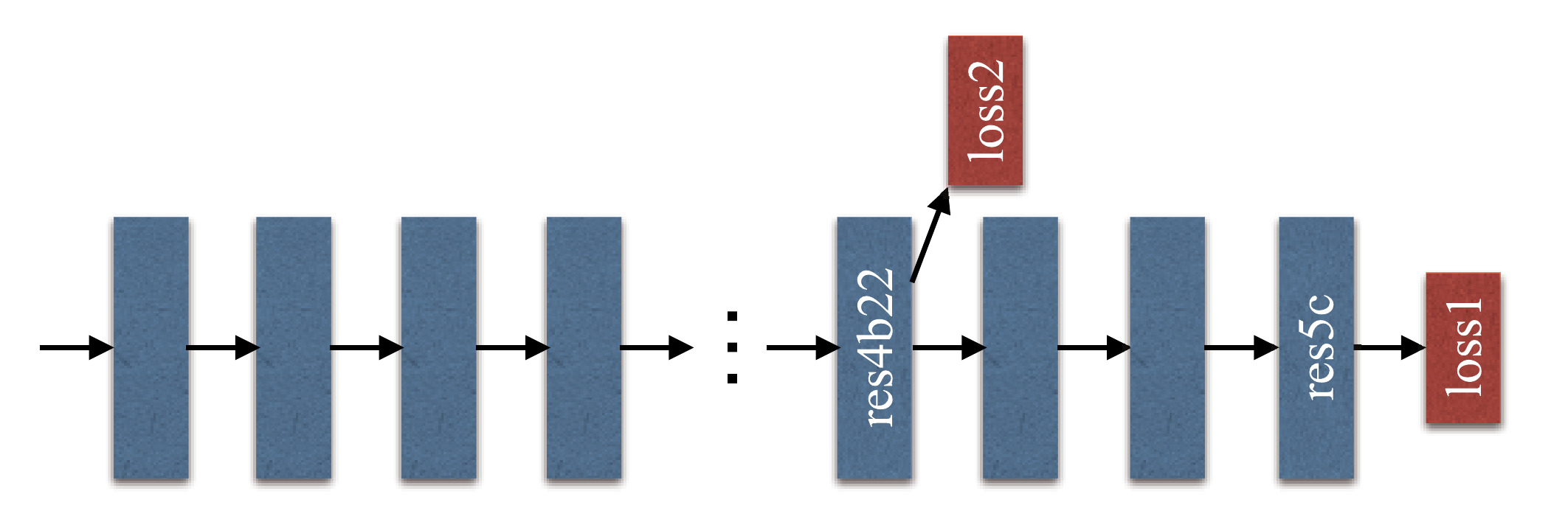}
\end{center}
\caption{Illustration of auxiliary loss in ResNet101. Each blue box denotes a residue
block. The auxiliary loss is added after the res4b22 residue block. }
\label{fig:auxiliaryloss}
\end{figure}

\section{Deep Supervision for ResNet-Based FCN}
Deep pretrained networks lead to good performance
\cite{krizhevsky2012imagenet,simonyan2014very,he2015deep}. However, increasing depth of
the network may introduce additional optimization difficulty as shown
in~\cite{shen2016relay,lee2015deeply} for image classification. ResNet solves this
problem with skip connection in each block. Latter layers of deep ResNet mainly learn
residues based on previous ones.

We contrarily propose generating initial results by supervision with an additional loss,
and learning the residue afterwards with the final loss. Thus, optimization of the deep
network is decomposed into two, each is simpler to solve.

An example of our deeply supervised ResNet101~\cite{he2015deep} model is illustrated in
Fig.~\ref{fig:auxiliaryloss}. Apart from the main branch using softmax loss to train the
final classifier, another classifier is applied after the fourth stage, i.e., the res4b22
residue block. Different from relay backpropagation~\cite{shen2016relay} that blocks the
backward auxiliary loss to several shallow layers, we let the two loss functions pass
through all previous layers. The auxiliary loss helps optimize the learning process,
while the master branch loss takes the most responsibility. We add weight to balance the
auxiliary loss.

In the testing phase, we abandon this auxiliary branch and only use the well optimized
master branch for final prediction. This kind of deeply supervised training strategy for
ResNet-based FCN is broadly useful under different experimental settings and works with
the pre-trained ResNet model. This manifests the generality of such a learning strategy.
More details are provided in Section~\ref{sec:imagenetexperimet}.

\section{Experiments}
Our proposed method is successful on scene parsing and semantic segmentation challenges.
We evaluate it in this section on three different datasets, including ImageNet scene
parsing challenge 2016~\cite{zhou2016semantic}, PASCAL VOC 2012 semantic
segmentation~\cite{everingham2010pascal} and urban scene understanding dataset
Cityscapes~\cite{cordts2016cityscapes}.

\subsection{Implementation Details}

For a practical deep learning system, devil is always in the details. Our implementation
is based on the public platform Caffe~\cite{jia2014caffe}. Inspired by
\cite{chen2016deeplab}, we use the ``poly" learning rate policy where current learning
rate equals to the base one multiplying $(1-\frac{iter}{max_iter})^{power}$. We set base
learning rate to 0.01 and power to 0.9. The performance can be improved by increasing the
iteration number, which is set to 150K for ImageNet experiment, 30K for PASCAL VOC and
90K for Cityscapes. Momentum and weight decay are set to 0.9 and 0.0001 respectively. For
data augmentation, we adopt random mirror and random resize between 0.5 and 2 for all
datasets, and additionally add random rotation between -10 and 10 degrees, and random
Gaussian blur for ImageNet and PASCAL VOC. This comprehensive data augmentation scheme
makes the network resist overfitting. Our network contains dilated convolution
following~\cite{chen2016deeplab}.

During the course of experiments, we notice that an appropriately large ``cropsize" can
yield good performance and ``batchsize" in the batch normalization~\cite{ioffe2015batch}
layer is of great importance. Due to limited physical memory on GPU cards, we set the
``batchsize" to 16 during training. To achieve this, we modify Caffe
from~\cite{wang2015towards} together with branch~\cite{chen2016deeplab} and make it
support batch normalization on data gathered from multiple GPUs based on OpenMPI. For the
auxiliary loss, we set the weight to 0.4 in experiments.

\subsection{ImageNet Scene Parsing Challenge 2016}
\label{sec:imagenetexperimet}

\paragraph{Dataset and Evaluation Metrics}
The ADE20K dataset \cite{zhou2016semantic} is used in ImageNet scene parsing challenge
2016. Different from other datasets, ADE20K is more challenging for the up to 150 classes
and diverse scenes with a total of 1,038 image-level labels. The challenge data is
divided into 20K/2K/3K images for training, validation and testing. Also, it needs to
parse both objects and stuff in the scene, which makes it more difficult than other
datasets. For evaluation, both {\it pixel-wise accuracy} (Pixel Acc.) and {\it mean of
class-wise intersection over union} (Mean IoU) are used.

\begin{table}
\footnotesize
\begin{center}
\begin{tabular}{ l c c}
\hline
Method & Mean IoU(\%) & Pixel Acc.(\%) \\
\hline\hline
ResNet50-Baseline & 37.23 & 78.01 \\
ResNet50+B1+MAX & 39.94 & 79.46 \\
ResNet50+B1+AVE & 40.07 & 79.52 \\
ResNet50+B1236+MAX & 40.18 & 79.45 \\
ResNet50+B1236+AVE & 41.07 & 79.97 \\
ResNet50+B1236+MAX+DR & 40.87 & 79.61 \\
ResNet50+B1236+AVE+DR & \textbf{41.68} & \textbf{80.04} \\
\hline
\end{tabular}
\end{center}
\caption{Investigation of PSPNet with different settings. Baseline is ResNet50-based FCN
with dilated network. `B1' and `B1236' denote pooled feature maps of bin sizes $\{1\times
1\}$ and $\{1\times1, 2\times2, 3\times3, 6\times6\}$ respectively. `MAX' and `AVE'
represent max pooling and average pooling operations individually. `DR' means that
dimension reduction is taken after pooling. The results are tested on the validation set
with the single-scale input.} \label{tab:pspnet}
\end{table}

\vspace{-0.1in}
\paragraph{Ablation Study for PSPNet}
To evaluate PSPNet, we conduct experiments with several settings, including pooling types
of max and average, pooling with just one global feature or four-level features, with and
without dimension reduction after the pooling operation and before concatenation. As
listed in Table \ref{tab:pspnet}, average pooling works better than max pooling in all
settings. Pooling with pyramid parsing outperforms that using global pooling. With
dimension reduction, the performance is further enhanced. With our proposed PSPNet, the
best setting yields results 41.68/80.04 in terms of Mean IoU and Pixel Acc. (\%), exceeding
global average pooling of 40.07/79.52 as idea in Liu~\etal~\cite{liu2015parsenet} by 1.61/0.52.
And compared to the baseline, PSPNet outperforming it by 4.45/2.03 in terms of absolute
improvement and 11.95/2.60 in terms of relative difference.

\begin{table}
\footnotesize
\begin{center}
\begin{tabular}{ l c c}
\hline
Loss Weight $\alpha$ & Mean IoU(\%) & Pixel Acc.(\%) \\
\hline\hline
ResNet50 (without AL) & 35.82 & 77.07 \\
ResNet50 (with $\alpha$ = 0.3) & 37.01 & 77.87 \\
ResNet50 (with $\alpha$ = 0.4) & \textbf{37.23} & \textbf{78.01} \\
ResNet50 (with $\alpha$ = 0.6) & 37.09 & 77.84 \\
ResNet50 (with $\alpha$ = 0.9) & 36.99 & 77.87 \\
\hline
\end{tabular}
\end{center}
\caption{Setting an appropriate loss weight $\alpha$ in the auxiliary branch is
important. `AL' denotes the auxiliary loss. Baseline is ResNet50-based FCN with dilated
network. Empirically, $\alpha$ = 0.4 yields the best performance. The results are tested
on the validation set with the single-scale input.} \label{tab:auxiliaryloss}
\end{table}

\begin{figure}[t]
\begin{center}
    \includegraphics[width=1.0\linewidth]{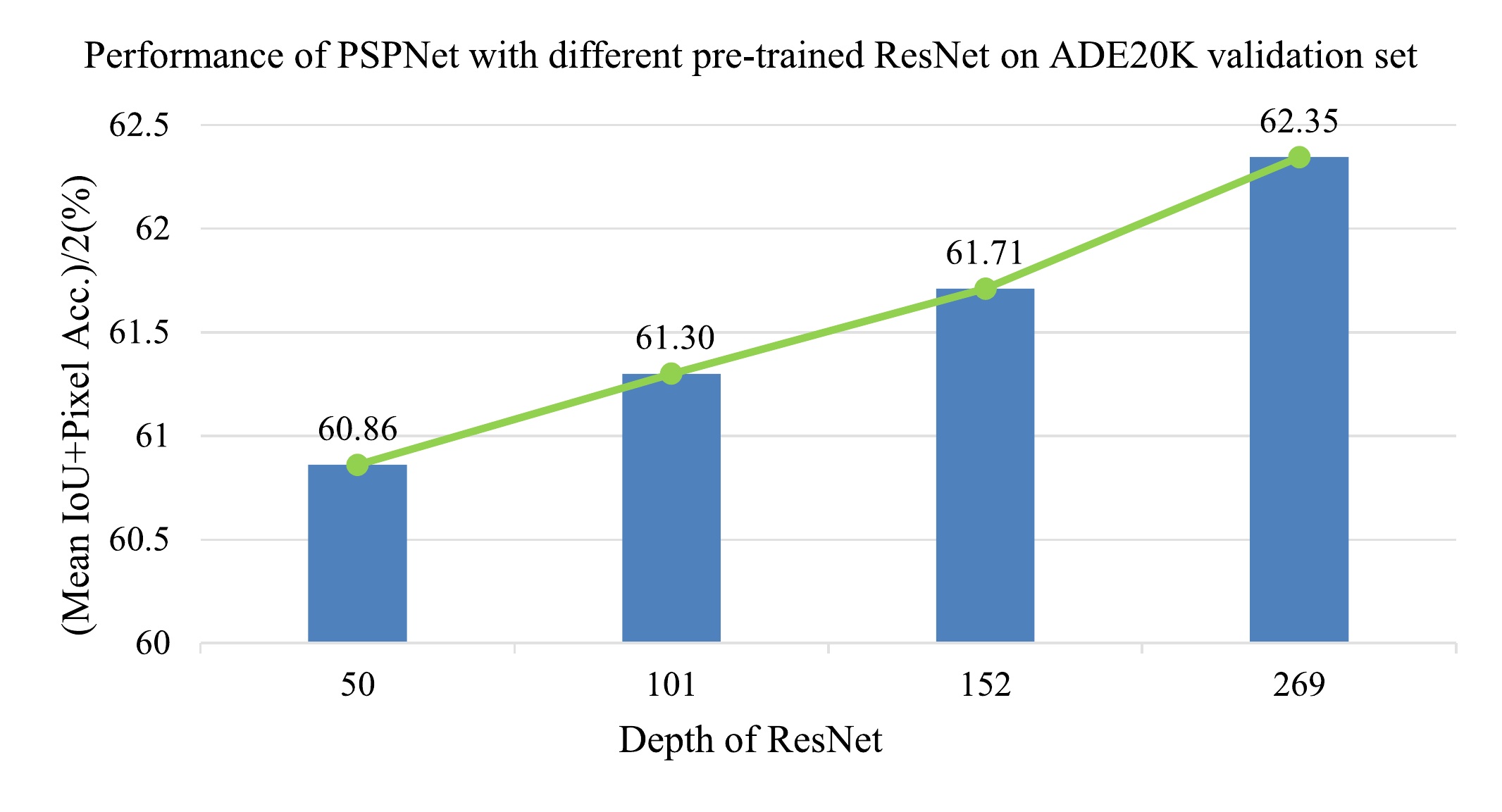}
\end{center}
\caption{Performance grows with deeper networks. The results are obtained on the
validation set with the single-scale input.} \label{fig:deepernet}
\end{figure}

\begin{table}
\footnotesize
\begin{center}
\begin{tabular}{ l c c}
\hline
Method & Mean IoU(\%) & Pixel Acc.(\%) \\
\hline\hline
PSPNet(50) & 41.68 & 80.04 \\
PSPNet(101) & 41.96 & 80.64 \\
PSPNet(152) & 42.62 & 80.80 \\
PSPNet(269) & \textbf{43.81} & \textbf{80.88} \\
\hline\hline
PSPNet(50)+MS & 42.78 & 80.76 \\
PSPNet(101)+MS & 43.29 & 81.39 \\
PSPNet(152)+MS & 43.51 & 81.38 \\
PSPNet(269)+MS & \textbf{44.94} & \textbf{81.69} \\
\hline
\end{tabular}
\end{center}
\caption{Deeper pre-trained model get higher performance. Number in the brackets refers
to the depth of ResNet and `MS' denotes multi-scale testing.} \label{tab:deepernet}
\end{table}

\begin{table}
\footnotesize
\begin{center}
\begin{tabular}{ l c c}
\hline
Method & Mean IoU(\%) & Pixel Acc.(\%) \\
\hline\hline
FCN \cite{long2015fully} & 29.39 & 71.32 \\
SegNet \cite{badrinarayanan2015segnet} & 21.64 & 71.00 \\
DilatedNet \cite{yu2015multi} & 32.31 & 73.55 \\
CascadeNet \cite{zhou2016semantic} & 34.90 & 74.52 \\
\hline
ResNet50-Baseline & 34.28 & 76.35 \\
ResNet50+DA & 35.82 & 77.07 \\
ResNet50+DA+AL & 37.23 & 78.01 \\
ResNet50+DA+AL+PSP & \textbf{41.68} & \textbf{80.04} \\
\hline
ResNet269+DA+AL+PSP & 43.81 & 80.88 \\
ResNet269+DA+AL+PSP+MS & \textbf{44.94} & \textbf{81.69} \\
\hline
\end{tabular}
\end{center}
\caption{Detailed analysis of our proposed PSPNet with comparison with others. Our
results are obtained on the validation set with the single-scale input except for the
last row. Results of FCN, SegNet and DilatedNet are reported in \cite{zhou2016semantic}.
`DA' refers to data augmentation we performed, `AL' denotes the auxiliary loss we added
and `PSP' represents the proposed PSPNet. `MS' means that multi-scale testing is used.}
\label{tab:baseline}
\end{table}

\begin{table}
\footnotesize
\begin{center}
\begin{tabular}{ c l c}
\hline
Rank & Team Name & Final Score (\%) \\
\hline\hline
\textbf{1} & \textbf{Ours} & \textbf{57.21} \\
2 & Adelaide & 56.74 \\
3 & 360+MCG-ICT-CAS\_SP & 55.56 \\
- & (our single model)  & (55.38) \\
4 & SegModel & 54.65 \\
5 & CASIA\_IVA & 54.33 \\
\hline
- & DilatedNet~\cite{yu2015multi} & 45.67 \\
- & FCN~\cite{long2015fully} & 44.80 \\
- & SegNet~\cite{badrinarayanan2015segnet} & 40.79 \\
\hline
\end{tabular}
\end{center}
\caption{Results of ImageNet scene parsing challenge 2016. The best entry of each team is
listed. The final score is the mean of Mean IoU and Pixel Acc. Results are evaluated on
the testing set. } \label{tab:imagenetscore}
\end{table}

\vspace{-0.1in}
\paragraph{Ablation Study for Auxiliary Loss}
The introduced auxiliary loss helps optimize the learning process while not influencing
learning in the master branch. We experiment with setting the auxiliary loss weight
$\alpha$ between 0 and 1 and show the results in Table~\ref{tab:auxiliaryloss}. The
baseline uses ResNet50-based FCN with dilated network, with the master branch's softmax
loss for optimization. Adding the auxiliary loss branch, $\alpha$ = 0.4 yields the best
performance. It outperforms the baseline with an improvement of 1.41/0.94 in terms of
Mean IoU and Pixel Acc. (\%). We believe deeper networks will benefit more given the new
augmented auxiliary loss.

\vspace{-0.1in}
\paragraph{Ablation Study for Pre-trained Model}
Deeper neural networks have been shown in previous work to be beneficial to large scale data
classification. To further analyze PSPNet, we conduct experiments for different depths of
pre-trained ResNet. We test four depths of \{50, 101, 152, 269\}. As shown in
Fig.~\ref{fig:deepernet}, with the same setting, increasing the depth of ResNet from 50
to 269 can improve the score of (Mean IoU + Pixel Acc.) / 2 (\%) from 60.86 to 62.35,
with 1.49 absolute improvement. Detailed scores of PSPNet pre-trained from different depth
ResNet models are listed in Table~\ref{tab:deepernet}.

\vspace{-0.1in}
\paragraph{More Detailed Performance Analysis}
We show our more detailed analysis on the validation set of ADE20K in Table
\ref{tab:baseline}. All our results except the last-row one use single-scale test.
``ResNet269+DA+AL+PSP+MS" uses multi-scale testing. Our baseline is adapted from
ResNet50 with dilated network, which yields MeanIoU 34.28 and Pixel Acc. 76.35. It
already outperforms other prior systems possibly due to the powerful
ResNet~\cite{he2015deep}.

Our proposed architecture makes further improvement compared to the baseline. Using data
augmentation, our result exceeds the baseline by 1.54/0.72 and reaches 35.82/77.07. Using
the auxiliary loss can further improve it by 1.41/0.94 and reaches 37.23/78.01. With
PSPNet, we notice relatively more significant progress for improvement of 4.45/2.03. The
result reaches 41.68/80.04. The difference from the baseline result is 7.40/3.69 in terms
of absolute improvement and 21.59/4.83 (\%) in terms of relativity. A deeper network of
ResNet269 yields even higher performance up to 43.81/80.88. Finally, the multi-scale
testing scheme moves the scores to 44.94/81.69.

\begin{figure}
\begin{center}
\includegraphics[width=1.0\linewidth]{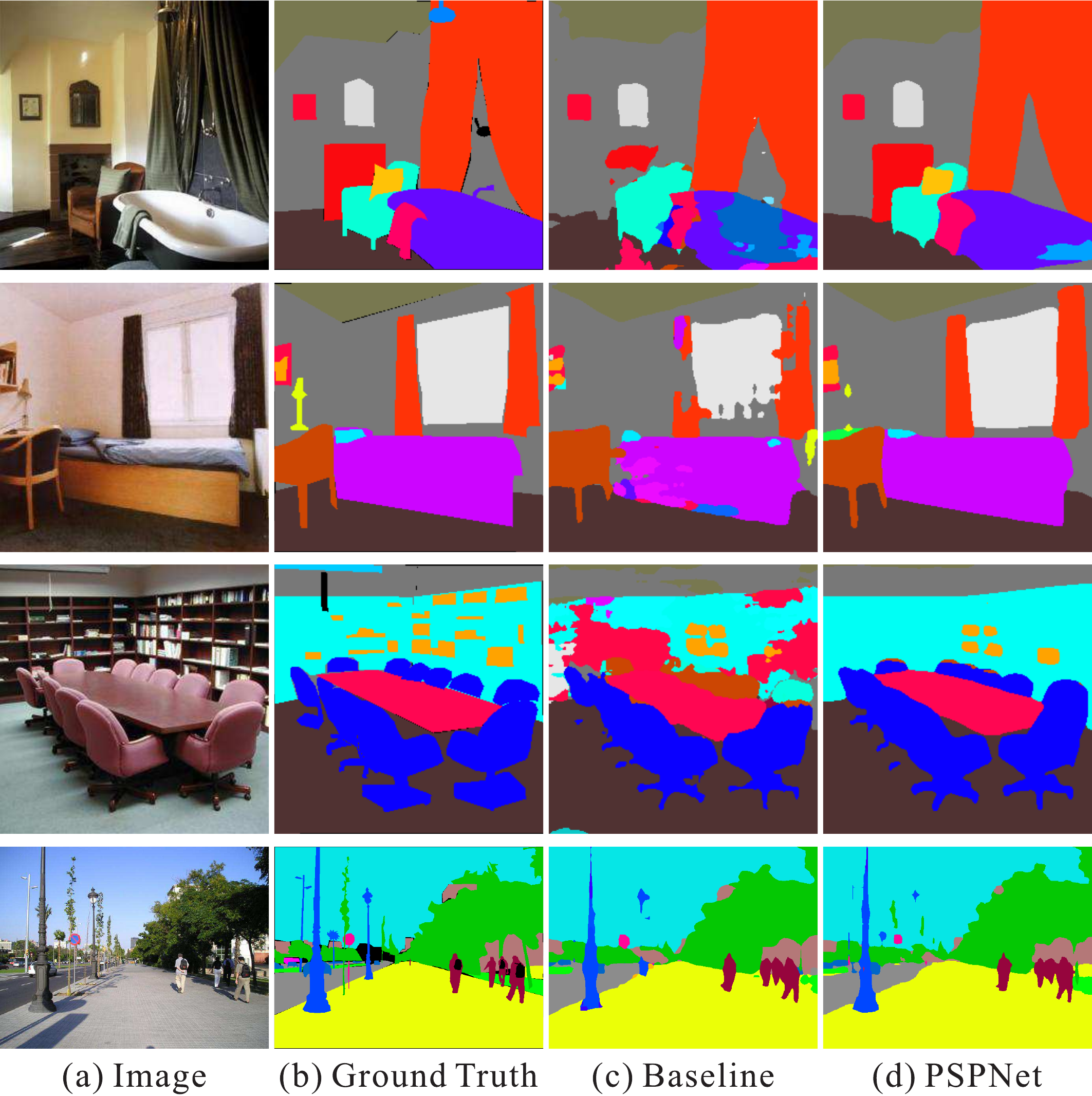}
\end{center}
\caption{Visual improvements on ADE20K, PSPNet produces more accurate and detailed
results.} \label{fig:imagenetimprove}
\end{figure}

\vspace{-0.1in}
\paragraph{Results in Challenge}
Using the proposed architecture, our team came in {the 1st place} in ImageNet scene
parsing challenge 2016. Table \ref{tab:imagenetscore} shows a few results in this competition.
Our ensemble submission achieves score 57.21\% on the testing set. Our single-model yields
score 55.38\%, which is even higher than a few other multi-model ensemble submissions.
This score is lower than that on the validation set possibly due to the difference of
data distributions between validation and testing sets. As shown in column (d) of Fig.
\ref{fig:fcnissues}, PSPNet solves the common problems in FCN.
Fig.~\ref{fig:imagenetimprove} shows another few parsing results on validation set of
ADE20K. Our results contain more accurate and detailed structures compared to the
baseline.

\begin{figure}
\begin{center}
\includegraphics[width=1.0\linewidth]{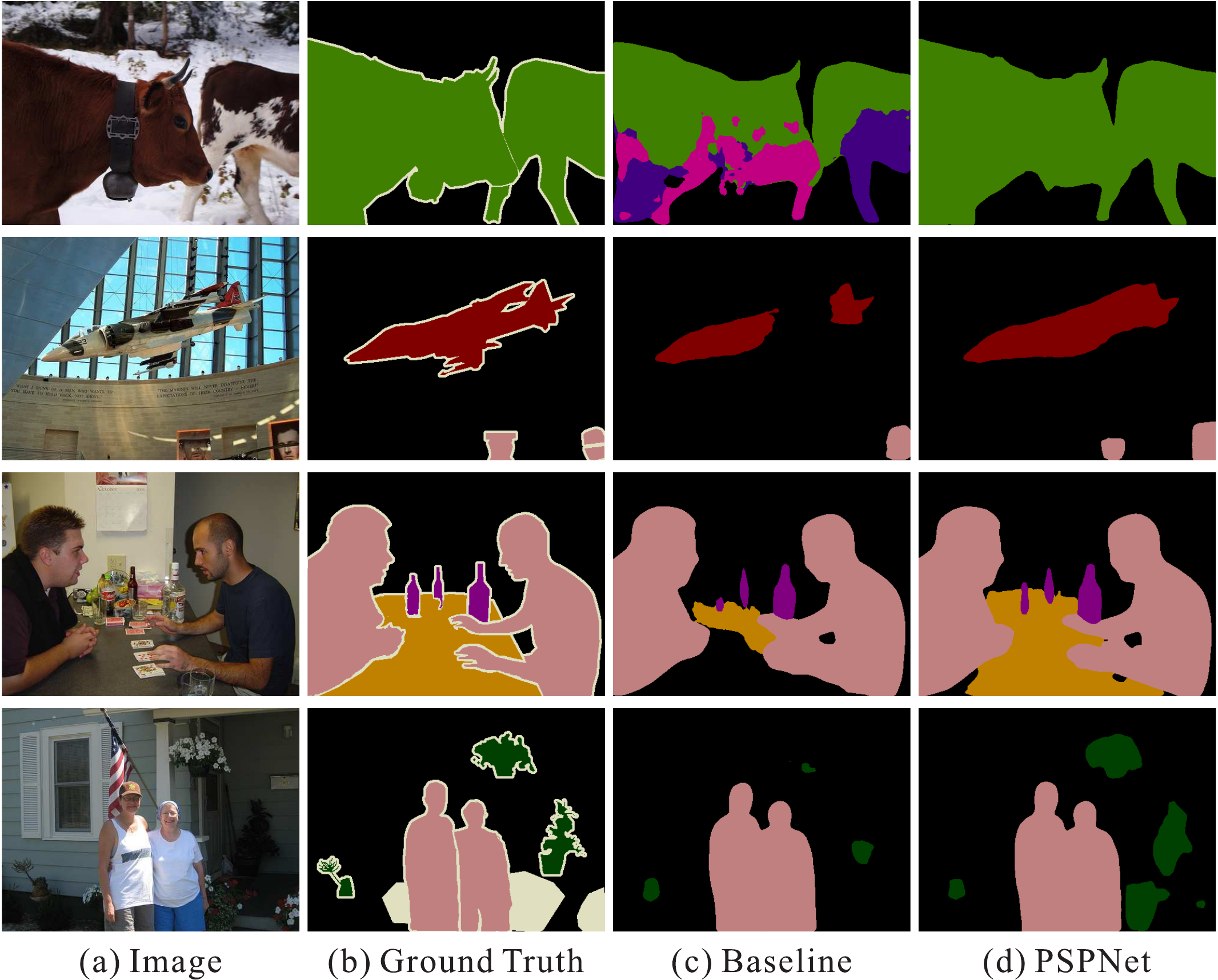}
\end{center}
\caption{Visual improvements on PASCAL VOC 2012 data. PSPNet produces more accurate and
detailed results.} \label{fig:vocimprove}
\end{figure}

\begin{table*}
    \footnotesize
    \setlength{\tabcolsep}{3pt}
    \begin{center}
        \begin{tabular}{ l | c c c c c c c c c c c c c c c c c c c c | c}
            \toprule[1pt]
            Method & aero & bike & bird & boat & bottle & bus & car & cat & chair & cow & table & dog & horse & mbike & person & plant & sheep & sofa & train & tv & mIoU \\
            \hline\hline
            FCN~\cite{long2015fully} & 76.8 & 34.2 & 68.9 & 49.4 & 60.3 & 75.3 & 74.7 & 77.6 & 21.4 & 62.5 & 46.8 & 71.8 & 63.9 & 76.5 & 73.9 & 45.2 & 72.4 & 37.4 & 70.9 & 55.1 & 62.2 \\
            Zoom-out~\cite{mostajabi2015feedforward} & 85.6 & 37.3 & 83.2 & 62.5 & 66.0 & 85.1 & 80.7 & 84.9 & 27.2 & 73.2 & 57.5 & 78.1 & 79.2 & 81.1 & 77.1 & 53.6 & 74.0 & 49.2 & 71.7 & 63.3 & 69.6 \\
            DeepLab~\cite{chen2014semantic} & 84.4 & 54.5 & 81.5 & 63.6 & 65.9 & 85.1 & 79.1 & 83.4 & 30.7 & 74.1 & 59.8 & 79.0 & 76.1 & 83.2 & 80.8 & 59.7 & 82.2 & 50.4 & 73.1 & 63.7 & 71.6 \\
            CRF-RNN~\cite{zheng2015conditional} & 87.5 & 39.0 & 79.7 & 64.2 & 68.3 & 87.6 & 80.8 & 84.4 & 30.4 & 78.2 & 60.4 & 80.5 & 77.8 & 83.1 & 80.6 & 59.5 & 82.8 & 47.8 & 78.3 & 67.1 & 72.0 \\
            DeconvNet~\cite{noh2015learning} & 89.9 & 39.3 & 79.7 & 63.9 & 68.2 & 87.4 & 81.2 & 86.1 & 28.5 & 77.0 & 62.0 & 79.0 & 80.3 & 83.6 & 80.2 & 58.8 & 83.4 & 54.3 & 80.7 & 65.0 & 72.5 \\
            GCRF~\cite{vemulapalli2016gaussian} & 85.2 & 43.9 & 83.3 & 65.2 & 68.3 & 89.0 & 82.7 & 85.3 & 31.1 & 79.5 & 63.3 & 80.5 & 79.3 & 85.5 & 81.0 & 60.5 & 85.5 & 52.0 & 77.3 & 65.1 & 73.2 \\
            DPN~\cite{liu2015semantic} & 87.7 & 59.4 & 78.4 & 64.9 & 70.3 & 89.3 & 83.5 & 86.1 & 31.7 & 79.9 & 62.6 & 81.9 & 80.0 & 83.5 & 82.3 & 60.5 & 83.2 & 53.4 & 77.9 & 65.0 & 74.1 \\
            Piecewise~\cite{lin2015efficient} & 90.6 & 37.6 & 80.0 & 67.8 & 74.4 & 92.0 & 85.2 & 86.2 & 39.1 & 81.2 & 58.9 & 83.8 & 83.9 & 84.3 & 84.8 & 62.1 & 83.2 & 58.2 & 80.8 & 72.3 & 75.3 \\
            PSPNet & \textbf{91.8} & \textbf{71.9} & \textbf{94.7} & \textbf{71.2} & \textbf{75.8} & \textbf{95.2} & \textbf{89.9} & \textbf{95.9} & \textbf{39.3} & \textbf{90.7} & \textbf{71.7} & \textbf{90.5} & \textbf{94.5} & \textbf{88.8} & \textbf{89.6} & \textbf{72.8} & \textbf{89.6} & \textbf{64.0} & \textbf{85.1} & \textbf{76.3} & \textbf{82.6} \\
            \hline\hline
            CRF-RNN$^\dag$~\cite{zheng2015conditional} & 90.4 & 55.3 & 88.7 & 68.4 & 69.8 & 88.3 & 82.4 & 85.1 & 32.6 & 78.5 & 64.4 & 79.6 & 81.9 & 86.4 & 81.8 & 58.6 & 82.4 & 53.5 & 77.4 & 70.1 & 74.7 \\
            BoxSup$^\dag$~\cite{dai2015boxsup} & 89.8 & 38.0 & 89.2 & 68.9 & 68.0 & 89.6 & 83.0 & 87.7 & 34.4 & 83.6 & 67.1 & 81.5 & 83.7 & 85.2 & 83.5 & 58.6 & 84.9 & 55.8 & 81.2 & 70.7 & 75.2 \\
            Dilation8$^\dag$~\cite{yu2015multi} & 91.7 & 39.6 & 87.8 & 63.1 & 71.8 & 89.7 & 82.9 & 89.8 & 37.2 & 84.0 & 63.0 & 83.3 & 89.0 & 83.8 & 85.1 & 56.8 & 87.6 & 56.0 & 80.2 & 64.7 & 75.3 \\
            DPN$^\dag$~\cite{liu2015semantic} & 89.0 & 61.6 & 87.7 & 66.8 & 74.7 & 91.2 & 84.3 & 87.6 & 36.5 & 86.3 & 66.1 & 84.4 & 87.8 & 85.6 & 85.4 & 63.6 & 87.3 & 61.3 & 79.4 & 66.4 & 77.5 \\
            Piecewise$^\dag$~\cite{lin2015efficient} & 94.1 & 40.7 & 84.1 & 67.8 & 75.9 & 93.4 & 84.3 & 88.4 & 42.5 & 86.4 & 64.7 & 85.4 & 89.0 & 85.8 & 86.0 & 67.5 & 90.2 & 63.8 & 80.9 & 73.0 & 78.0 \\
            FCRNs$^\dag$~\cite{wu2016bridging} & 91.9 & 48.1 & 93.4 & 69.3 & 75.5 & 94.2 & 87.5 & 92.8 & 36.7 & 86.9 & 65.2 & 89.1 & 90.2 & 86.5 & 87.2 & 64.6 & 90.1 & 59.7 & 85.5 & 72.7 & 79.1 \\
            LRR$^\dag$~\cite{ghiasi2016laplacian} & 92.4 & 45.1 & 94.6 & 65.2 & 75.8 & \textbf{95.1} & 89.1 & 92.3 & 39.0 & 85.7 & 70.4 & 88.6 & 89.4 & 88.6 & 86.6 & 65.8 & 86.2 & 57.4 & 85.7 & 77.3 & 79.3 \\
            DeepLab$^\dag$~\cite{chen2016deeplab} & 92.6 & 60.4 & 91.6 & 63.4 & 76.3 & 95.0 & 88.4 & 92.6 & 32.7 & 88.5 & 67.6 & 89.6 & 92.1 & 87.0 & 87.4 & 63.3 & 88.3 & 60.0 & 86.8 & 74.5 & 79.7 \\
            PSPNet$^\dag$ & \textbf{95.8} & \textbf{72.7} & \textbf{95.0} & \textbf{78.9} & \textbf{84.4} & 94.7 & \textbf{92.0} & \textbf{95.7} & \textbf{43.1} & \textbf{91.0} & \textbf{80.3} & \textbf{91.3} & \textbf{96.3} & \textbf{92.3} & \textbf{90.1} & \textbf{71.5} & \textbf{94.4} & \textbf{66.9} & \textbf{88.8} & \textbf{82.0} & \textbf{85.4} \\
            \bottomrule[1pt]
        \end{tabular}
    \end{center}
    \caption{Per-class results on PASCAL VOC 2012 testing set. Methods pre-trained on MS-COCO
        are marked with `\dag'.} \label{tab:vocresult}
\end{table*}

\subsection{PASCAL VOC 2012}\label{sec:vocexperiment}
Our PSPNet also works satisfyingly on semantic segmentation. We carry out experiments on
the PASCAL VOC 2012 segmentation dataset~\cite{everingham2010pascal}, which contains 20
object categories and one background class. Following the procedure of
\cite{long2015fully,dai2015boxsup,papandreou2015weakly,chen2014semantic}, we use
augmented data with the annotation of~\cite{hariharan2011semantic} resulting 10,582,
1,449 and 1,456 images for training, validation and testing. Results are shown in
Table~\ref{tab:vocresult}, we compare PSPNet with previous best-performing methods on the
testing set based on two settings, i.e., with or without pre-training on MS-COCO
dataset~\cite{lin2014microsoft}. Methods pre-trained with MS-COCO are marked by `\dag'.
For fair comparison with current ResNet based frameworks~\cite{wu2016bridging,ghiasi2016laplacian,chen2016deeplab}
in scene parsing/semantic segmentation task, we build our architecture
based on ResNet101 while without post-processing like CRF. We evaluate PSPNet with several-scale input and use the average results
following~\cite{chen2014semantic,liu2015parsenet}.

As shown in Table~\ref{tab:vocresult}, PSPNet outperforms prior methods on both settings.
Trained with only VOC 2012 data, we achieve 82.6\%
accuracy\footnote{\href{http://host.robots.ox.ac.uk:8080/anonymous/0OOWLP.html}{http://host.robots.ox.ac.uk:8080/anonymous/0OOWLP.html}}
-- we get the highest accuracy on all 20 classes. When PSPNet is pre-trained with MS-COCO
dataset, it reaches 85.4\%
accuracy\footnote{\href{http://host.robots.ox.ac.uk:8080/anonymous/6KIR41.html}{http://host.robots.ox.ac.uk:8080/anonymous/6KIR41.html}}
where 19 out of the 20 classes receive the highest accuracy. Intriguingly, our PSPNet
trained with only VOC 2012 data outperforms existing methods trained with the MS-COCO
pre-trained model.

One may argue that our based classification model is more powerful than several prior methods
since ResNet was recently proposed. To exhibit our unique contribution, we show that our
method also outperforms state-of-the-art frameworks that use the same model, including
FCRNs~\cite{wu2016bridging}, LRR~\cite{ghiasi2016laplacian}, and
DeepLab~\cite{chen2016deeplab}. In this process, we even do not employ time-consuming but
effective post-processing, such as CRF, as that in~\cite{chen2016deeplab,ghiasi2016laplacian}.

\begin{figure*}
	\begin{center}
		\includegraphics[width=0.93\linewidth]{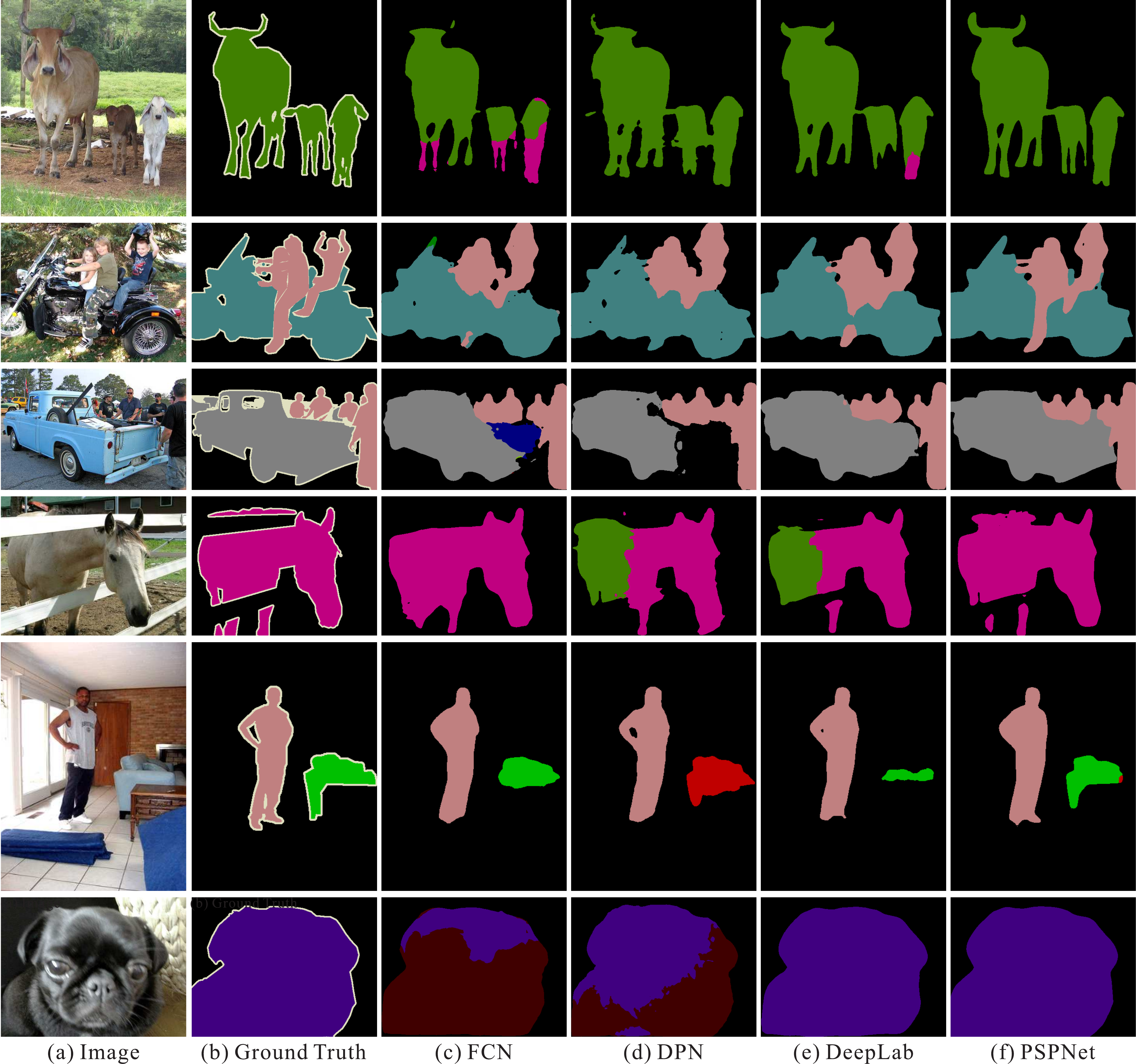}
	\end{center}
	\caption{Visual comparison on PASCAL VOC 2012 data. (a) Image. (b) Ground Truth.
		(c) FCN~\cite{long2015fully}. (d) DPN~\cite{liu2015parsenet}. (e) DeepLab~\cite{chen2016deeplab}. (f) PSPNet.}
	\label{fig:vocresult}
\end{figure*}

Several examples are shown in Fig.~\ref{fig:vocimprove}. For ``cows" in row one, our
baseline model treats it as ``horse" and ``dog" while PSPNet corrects these errors. For
``aeroplane" and ``table" in the second and third rows, PSPNet finds missing parts. For
``person", ``bottle" and ``plant" in following rows, PSPNet performs well on these
small-size-object classes in the images compared to the baseline model. More visual
comparisons between PSPNet and other methods are included in Fig.~\ref{fig:vocresult}.

\begin{table}
    \footnotesize
    \begin{center}
        \begin{tabular}{ l | c c c c}
            \toprule[1pt]
            Method & IoU cla. & iIoU cla. & IoU cat. & iIoU cat. \\
            \hline\hline
            CRF-RNN~\cite{zheng2015conditional} & 62.5 & 34.4 & 82.7 & 66.0 \\
            FCN~\cite{long2015fully} & 65.3 & 41.7 & 85.7 & 70.1 \\
            SiCNN~~\cite{krevso2016convolutional} & 66.3 & 44.9 & 85.0 & 71.2 \\
            DPN~\cite{liu2015semantic} & 66.8 & 39.1 & 86.0 & 69.1 \\
            Dilation10~\cite{yu2015multi} & 67.1 & 42.0 & 86.5 & 71.1 \\
            LRR~\cite{ghiasi2016laplacian} & 69.7 & 48.0 & 88.2 & 74.7 \\
            DeepLab~\cite{chen2016deeplab} & 70.4 & 42.6 & 86.4 & 67.7 \\
            Piecewise~\cite{lin2015efficient} & 71.6 & 51.7 & 87.3 & 74.1 \\
            PSPNet & \textbf{78.4} & \textbf{56.7} & \textbf{90.6} & \textbf{78.6} \\
            \hline\hline
            LRR$^\ddag$~\cite{ghiasi2016laplacian} & 71.8 & 47.9 & 88.4 & 73.9 \\
            PSPNet$^\ddag$ & \textbf{80.2} & \textbf{58.1} & \textbf{90.6} & \textbf{78.2} \\
            \bottomrule[1pt]
        \end{tabular}
    \end{center}
    \caption{Results on Cityscapes testing set. Methods trained using both fine and
        coarse data are marked with `$\ddag$'.}
    \label{tab:cityscapescomprison}
\end{table}

\begin{figure}
\begin{center}
\includegraphics[width=0.95\linewidth]{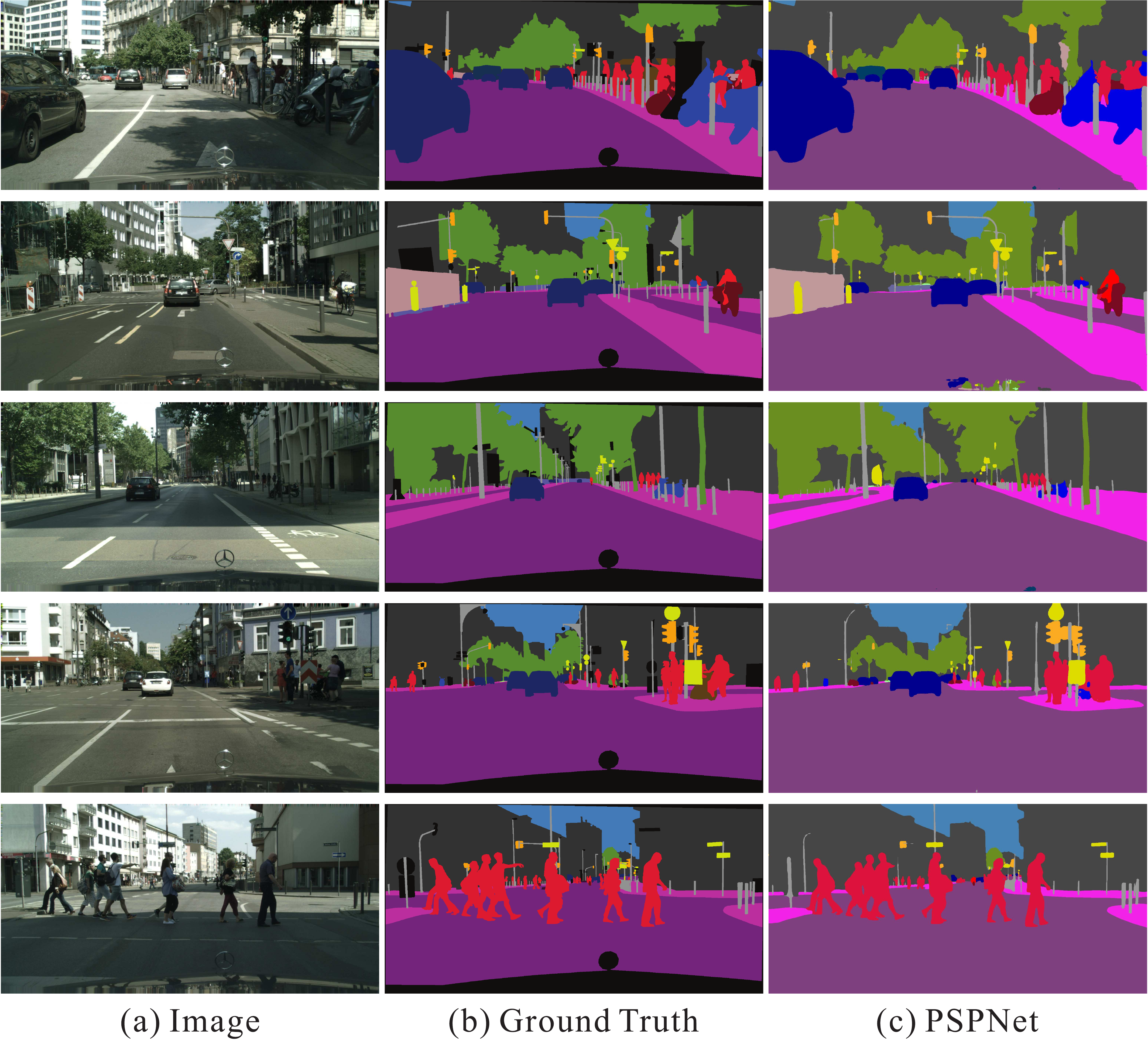}
\end{center}
   \caption{Examples of PSPNet results on Cityscapes dataset.}
\label{fig:cityscapesresult}
\end{figure}

\subsection{Cityscapes}
Cityscapes~\cite{cordts2016cityscapes} is a recently released dataset for semantic urban
scene understanding. It contains 5,000 high quality pixel-level finely annotated images
collected from 50 cities in different seasons. The images are divided into sets with
numbers 2,975, 500, and 1,525 for training, validation and testing. It defines 19
categories containing both stuff and objects. Also, 20,000 coarsely annotated images are
provided for two settings in comparison, i.e., training with only fine data or with both
the fine and coarse data. Methods trained using both fine and coarse data are marked with
`$\ddag$'. Detailed results are listed in Table~\ref{tab:cityscapescomprison}. Our base model is
ResNet101 as in DeepLab~\cite{chen2016deeplab} for fair comparison and the testing procedure
follows Section~\ref{sec:vocexperiment}.

Statistics in Table~\ref{tab:cityscapescomprison} show that PSPNet outperforms other
methods with notable advantage. Using both fine and coarse data for training makes our
method yield 80.2 accuracy. Several examples are shown in
Fig.~\ref{fig:cityscapesresult}. Detailed per-class results on testing set are shown in
Table~\ref{tab:cityscapesresult}.

\vspace{-0.05in}
\section{Concluding Remarks}
We have proposed an effective pyramid scene parsing network for complex scene
understanding. The global pyramid pooling feature provides additional contextual
information. We have also provided a deeply supervised optimization strategy for
ResNet-based FCN network.
We hope the implementation details publicly available can help
the community adopt these useful strategies for scene parsing and semantic segmentation
and advance related techniques.

\section*{Acknowledgements}

We would like to thank Gang Sun and Tong Xiao for their help in training the basic
classification models, Qun Luo for technical support. This work is supported by a grant from the Research Grants Council
of the Hong Kong SAR (project No. 2150760).

\begin{table*}[t]
	\footnotesize
	\setlength{\tabcolsep}{3.1pt}
	\begin{center}
		\begin{tabular}{ l | c c c c c c c c c c c c c c c c c c c | c}
			\toprule[1pt]
			Method & road & swalk & build. & wall & fence & pole & tlight & sign & veg. & terrain & sky & person & rider & car & truck & bus & train & mbike & bike & mIoU \\
			\hline\hline
			CRF-RNN~\cite{zheng2015conditional} & 96.3 & 73.9 & 88.2 & 47.6 & 41.3 & 35.2 & 49.5 & 59.7 & 90.6 & 66.1 & 93.5 & 70.4 & 34.7 & 90.1 & 39.2 & 57.5 & 55.4 & 43.9 & 54.6 & 62.5 \\
			FCN~\cite{long2015fully} & 97.4 & 78.4 & 89.2 & 34.9 & 44.2 & 47.4 & 60.1 & 65.0 & 91.4 & 69.3 & 93.9 & 77.1 & 51.4 & 92.6 & 35.3 & 48.6 & 46.5 & 51.6 & 66.8 & 65.3 \\
			SiCNN+CRF~\cite{krevso2016convolutional} & 96.3 & 76.8 & 88.8 & 40.0 & 45.4 & 50.1 & 63.3 & 69.6 & 90.6 & 67.1 & 92.2 & 77.6 & 55.9 & 90.1 & 39.2 & 51.3 & 44.4 & 54.4 & 66.1 & 66.3 \\
			DPN~\cite{liu2015semantic} & 97.5 & 78.5 & 89.5 & 40.4 & 45.9 & 51.1 & 56.8 & 65.3 & 91.5 & 69.4 & 94.5 & 77.5 & 54.2 & 92.5 & 44.5 & 53.4 & 49.9 & 52.1 & 64.8 & 66.8 \\
			Dilation10~\cite{yu2015multi} & 97.6 & 79.2 & 89.9 & 37.3 & 47.6 & 53.2 & 58.6 & 65.2 & 91.8 & 69.4 & 93.7 & 78.9 & 55.0 & 93.3 & 45.5 & 53.4 & 47.7 & 52.2 & 66.0 & 67.1 \\
			LRR~\cite{ghiasi2016laplacian} & 97.7 & 79.9 & 90.7 & 44.4 & 48.6 & 58.6 & 68.2 & 72.0 & 92.5 & 69.3 & 94.7 & 81.6 & 60.0 & 94.0 & 43.6 & 56.8 & 47.2 & 54.8 & 69.7 & 69.7 \\
			DeepLab~\cite{chen2016deeplab} & 97.9 & 81.3 & 90.3 & 48.8 & 47.4 & 49.6 & 57.9 & 67.3 & 91.9 & 69.4 & 94.2 & 79.8 & 59.8 & 93.7 & 56.5 & 67.5 & 57.5 & 57.7 & 68.8 & 70.4 \\
			Piecewise~\cite{lin2015efficient} & 98.0 & 82.6 & 90.6 & 44.0 & 50.7 & 51.1 & 65.0 & 71.7 & 92.0 & 72.0 & 94.1 & 81.5 & 61.1 & 94.3 & 61.1 & 65.1 & 53.8 & 61.6 & 70.6 & 71.6 \\
			PSPNet & \textbf{98.6} & \textbf{86.2} & \textbf{92.9} & \textbf{50.8} & \textbf{58.8} & \textbf{64.0} & \textbf{75.6} & \textbf{79.0} & \textbf{93.4} & \textbf{72.3} & \textbf{95.4} & \textbf{86.5} & \textbf{71.3} & \textbf{95.9} & \textbf{68.2} & \textbf{79.5} & \textbf{73.8} & \textbf{69.5} & \textbf{77.2} & \textbf{78.4}\\
			\hline\hline
			LRR$^\ddag$~\cite{ghiasi2016laplacian} & 97.9 & 81.5 & 91.4 & 50.5 & 52.7 & 59.4 & 66.8 & 72.7 & 92.5 & 70.1 & 95.0 & 81.3 & 60.1 & 94.3 & 51.2 & 67.7 & 54.6 & 55.6 & 69.6 & 71.8 \\
			PSPNet$^\ddag$ & \textbf{98.6} & \textbf{86.6} & \textbf{93.2} & \textbf{58.1} & \textbf{63.0} & \textbf{64.5} & \textbf{75.2} & \textbf{79.2} & \textbf{93.4} & \textbf{72.1} & \textbf{95.1} & \textbf{86.3} & \textbf{71.4} & \textbf{96.0} & \textbf{73.5} & \textbf{90.4} & \textbf{80.3} & \textbf{69.9} & \textbf{76.9} & \textbf{80.2} \\
			\bottomrule[1pt]
		\end{tabular}
	\end{center}
	\caption{Per-class results on Cityscapes testing set. Methods trained using both fine and coarse set are marked with `$\ddag$'.}
	\label{tab:cityscapesresult}
\end{table*}

{\small
\bibliographystyle{ieee}
\bibliography{egbib}
}

\end{document}